\documentclass[letterpaper, 10 pt, conference]{ieeeconf}  

\IEEEoverridecommandlockouts                              

\usepackage{amsmath} 
\usepackage{amsfonts}
\usepackage{soul,color}
\usepackage{booktabs}
\usepackage{graphicx}
\usepackage{lipsum} 
\usepackage{algorithm}
\usepackage{algpseudocode}
\usepackage{gensymb}
\usepackage{cite}
\usepackage{multirow}

\title{\LARGE \bf
Topologically Persistent Features-based Object Recognition in Cluttered Indoor Environments
}

\author{Ekta U. Samani$^{1}$ and Ashis G. Banerjee$^{2}$
\thanks{$^{1}$E. U. Samani is with the Department of Mechanical Engineering, University of Washington, Seattle, WA 98195, USA,
        {\tt\small ektas@uw.edu}}%
\thanks{$^{2}$A. G. Banerjee is with the Department of Industrial \& Systems Engineering and the Department of Mechanical Engineering, University of Washington, Seattle, WA 98195, USA,
        {\tt\small ashisb@uw.edu}}%
}

\begin{document}

\maketitle
\thispagestyle{empty}
\pagestyle{empty}




\begin{abstract}



Recognition of occluded objects in unseen indoor environments is a challenging problem for mobile robots. This work proposes a new slicing-based topological descriptor that captures the 3D shape of object point clouds to address this challenge. 
It yields similarities between the descriptors of the occluded and the corresponding unoccluded objects, enabling object unity-based recognition using a library of trained models. The descriptor is obtained by 
partitioning an object's point cloud into multiple 2D slices 
and constructing filtrations (nested sequences of simplicial complexes) on the slices to mimic further slicing of the slices, thereby capturing detailed shapes through persistent homology-generated features.
We use nine different sequences of cluttered scenes from a benchmark dataset for performance evaluation. Our method outperforms two state-of-the-art deep learning-based point cloud classification methods, namely, DGCNN and SimpleView.

\end{abstract}

\section{Introduction}

Object recognition is crucial for most robot vision systems to obtain a semantic-level understanding of the robot's environment. Early deep learning methods achieved extraordinary performance in this task. However, these methods are sensitive to variations in 
illumination, background, and object appearance. Consequently, they are insufficient in dealing with the challenges associated with long-term robot autonomy, where the robots operate in complex and continually-changing environments for extended time periods. Efforts have been made to address this challenge by developing domain adaptation methods for cross-domain object detection using adversarial learning \cite{chen2018domain,he2019multi, saito2019strong, kim2019diversify}. Alternatively, in our previous work \cite{samani2021visual}, we investigated the use of domain-invariant, topologically persistent features, which capture the shape information to achieve robust object recognition. The framework showed more robustness to environmental variations than a state-of-the-art method. However, it only uses the objects' 2D shape information and encounters difficulties when the shape changes considerably due to large variations in the camera pose and object occlusions.

Recognition using features that capture the 3D shape of the objects can alleviate the difficulty associated with camera pose variations. Several geometric and topological descriptors for 3D point clouds of objects have been proposed in the literature \cite{rusu2009detecting,marton2010hierarchical, aldoma2012our, wohlkinger2011ensemble, aldoma2011cad, guo20143d,biasotti2016recent,beksi2018signature}. However, they are unsuitable when the point clouds are incomplete due to partial occlusion. Deep learning-based methods have also been proposed to obtain features of geometric data \cite{qi2017pointnet,qi2017pointnet++,wang2019dynamic,goyal2021revisiting}. However, such features also run into robustness issues when the point clouds are corrupted due to occlusion \cite{ren2022benchmarking}. Therefore, we propose a new 
approach for computing topologically persistent features that capture the innate 3D shape information from the point clouds 
to recognize occluded objects.


The key contributions of our work are as follows:
\begin{itemize}

     \item We develop a new slicing-based descriptor function to capture the detailed shape of objects while ensuring similarities in the descriptors of the occluded objects and the corresponding unoccluded objects.
     

    \item We propose a recognition framework aligned with visual object recognition in humans \cite{rapp2015handbook}, based on the idea of \textit{object unity}, a human reasoning mechanism \cite{goldstein2016sensation}. 
    \item We show that our method outperforms two state-of-the-art deep learning-based methods for point cloud classification in cluttered environments with a wide variety of objects.
\end{itemize}

\section{Mathematical Preliminaries}
In topological data analysis (TDA), a point cloud is commonly represented using a simplicial complex. A simplicial complex $K$ is a finite union of simplices in $\mathbb{R}^n$ such that every face of a simplex from $K$ is also in $K$, and, the non-empty intersection of any two simplices in $K$ is a face of both the simplices. Persistent homology is applied to compute the topological features of point clouds using such complexes. To compute the features, a filtration of simplicial complexes is constructed from the point cloud. A filtration is a nested sequence of complexes $K_{1}, \ldots, K_{r}$ such that $ K_{1}\subseteq . . . \subseteq K_{r} = K$. A common way to generate such a filtration is to consider the sublevel sets $K_\mathtt{t} = f_d^{-1} ([-\infty, \mathtt{t}])$ of a descriptor function $f_d : \mathbb{X} \longrightarrow \mathbb{R}$ on a topological space $\mathbb{X}$ indexed by a parameter $\mathtt{t} \in \mathbb{R}$. As $\mathtt{t}$ increases from $(-\infty,\infty)$, topological features appear and disappear in the filtration, referred to as their \textit{birth} and \textit{death} times, respectively. This topological information is summarized in an $m$-dimensional persistence diagram (PD). An $m$-dimensional PD is a countable multiset of points in $\mathbb{R}^{2}$. Each point $(\mathtt{x},\mathtt{y})$ represents an $m$-dimensional hole born at a time $\mathtt{x}$ and filled at a time $\mathtt{y}$. The diagonal of a PD is a multiset $\Delta = \left\{ (\mathtt{x},\mathtt{x}) \in \mathbb{R}^{2} \vert \mathtt{x} \in \mathbb{R}\right\}$, where every point in $\Delta$ has infinite multiplicity. A persistence image (PI) is a stable and finite dimensional vector representation generated from a PD \cite{adams2017persistence}. To obtain a PI, an equivalent diagram of birth-persistence points, i.e., $(\mathtt{x},\mathtt{y}-\mathtt{x})$, is computed. These points are then regarded as a sum of Dirac delta functions, which are convolved with a Gaussian kernel over a rectangular grid of evenly sampled points to compute the PI.

\section{Method}
Given a real-world RGB-D image of a cluttered scene and the corresponding instance segmentation map, our goal is to recognize all the objects in the scene based on their 3D shape information. First, we generate the individual point clouds of all the objects in the scene using the depth image. We then perform view normalization \cite{rapp2015handbook} to suitably align every point cloud and compute its slicing-based topological descriptor. We then perform recognition using a library of classifiers trained using synthetic images. Fig. \ref{pipeline} shows the proposed framework. The following subsections describe the steps in the framework.

\subsection{View normalization}

Consider an object point cloud $\mathcal{P}$ in $\mathbb{R}^{3}$. To obtain the desired alignment, we first compute the minimal volume bounding box of $\mathcal{P}$ using a principal components analysis (PCA)-based approximation of the O'Rourke's algorithm \cite{o1985finding}. The bounding box is oriented such that the coordinate axes are ordered with respect to the principal components. We then rotate the point cloud such that the minimal volume bounding box of the rotated point cloud is aligned with the coordinate axes. 
We then perform translation such that the resultant point cloud lies in the first octant. Last, the point cloud is rotated by an angle $\alpha$ about the $y$-axis to obtain $\hat{\mathcal{P}}$. 

\subsection{Slicing-based descriptor generation}
To obtain a slicing-based descriptor of $\hat{\mathcal{P}}$, first, 
we slice $\hat{\mathcal{P}}$ along the $z$-axis to get slices $\mathcal{S}^i$, where $i \in \mathbb{Z} \cap [0,\frac{h}{\sigma_1}]$. Here, $h$ is the dimension of the axis-aligned bounding box of $\hat{\mathcal{P}}$ along the $z$-axis, and $\sigma_1$ represents is the desired 'thickness' of the slices. Let $p=(p_x,p_y,p_z)$ be a point in $\hat{\mathcal{P}}$. The slices $\mathcal{S}^i$ can be obtained as follows. 
\begin{equation}
\mathcal{S}^{i}:=\left\{p \in \hat{\mathcal{P}} \mid i\sigma_1 \leq p_z < (i+1)\sigma_1    \right\}.
\end{equation}
Let $s=(s_x,s_y,s_z)$ represent a point in $\mathcal{S}^i$. For every slice $\mathcal{S}^i$, we modify the $z$-coordinates $\forall s \in \mathcal{S}^i$ to $s_z^\prime$, where $s_z^\prime=i\sigma_1$.

We then design a descriptor function to build a filtration from every slice. The function is designed to mimic further slicing of the slice along the $x$-axis, thereby capturing the shape of the slice in the PD. To obtain the PDs, first, we compute a set of origin points, $\mathcal{O}^i$, and a set of termination points, $\mathcal{T}^i$, for every slice $\mathcal{S}^i$. Let $o=(o_x,o_y,o_z)$ and $t=(t_x,t_y,t_z)$ represent a point in $\mathcal{O}^i$ and $\mathcal{T}^i$, respectively. The sets $\mathcal{O}^i$ and $\mathcal{T}^i$ can be constructed as follows.
\begin{equation}
\label{origin}
\begin{array}{lcl}
\mathcal{O}^i &:=& \left\{o, \forall j \in \mathbb{Z} \cap [0,\frac{w}{\sigma_2}] \mid  o_x = (j+1)\sigma_2,\right. \\
&&\left.  o_y = \inf(\{s_y \forall s \in \mathcal{S}^i \mid j\sigma_2 \leq s_x < (j+1)\sigma_2\}),\right.\\
&& \left. o_z = i\sigma_1 + \epsilon_1 \right\},
\end{array}
\end{equation}

\begin{equation}
\label{termination}
\begin{array}{lcl}
\mathcal{T}^i &:=& \left\{t, \forall j \in \mathbb{Z} \cap [0,\frac{w}{\sigma_2}] \mid  t_x = (j+1)\sigma_2,\right. \\
&&\left.  t_y = \sup(\{s_y \forall s \in \mathcal{S}^i \mid j\sigma_2 \leq s_x < (j+1)\sigma_2\}),\right.\\
&& \left. t_z = i\sigma_1 + \epsilon_2 \right\},
\end{array}
\end{equation}
where $w$ is the dimension of the axis-aligned bounding box of $\mathcal{S}^i$ along the $x$-axis, $\sigma_2$ represents the desired 'thickness' of a slice if further slicing of $\mathcal{S}^i$ is performed along the $x$-axis, and $\epsilon_1,\epsilon_2$ are arbitrarily small positive constants ($2\epsilon_1 < \epsilon_2$). For every slice $\mathcal{S}^i$, we then modify the $x$-coordinates $\forall s \in \mathcal{S}^i$ to $s_x^\prime$ such that if $j\sigma_2 \leq s_x < (j+1)\sigma_2$, then $s_x^\prime = (j+1)\sigma_2$.

The descriptor function, $f$, to construct a filtration from every $\mathcal{S}^i$ is then defined as 
\begin{equation}
\label{filtouter}
  f(a,b) =
    \begin{cases}
      0 & \text{if  $a,b \in \mathcal{T}^i$  }\\
      g(a,b) & \text{otherwise},
    \end{cases}       
\end{equation}
where $a=(a_x,a_y,a_z)$ and $b=(b_x,b_y,b_z)$ are any two points in $\mathcal{S}^i \cup \mathcal{O}^i \cup \mathcal{T}^i$. The function $g$ is computed as follows.
\begin{equation}
\label{filtinner}
  g(a,b) =
    \begin{cases}
      \infty & \text{if $a_x \neq b_x$ or $|a_z - b_z| = \epsilon_2$}  \\ 

      a_x + |a_y - b_y| & \text{otherwise}.
    \end{cases}  
\end{equation}

\begin{figure}[b]
    \centering
    \includegraphics[width=\columnwidth]{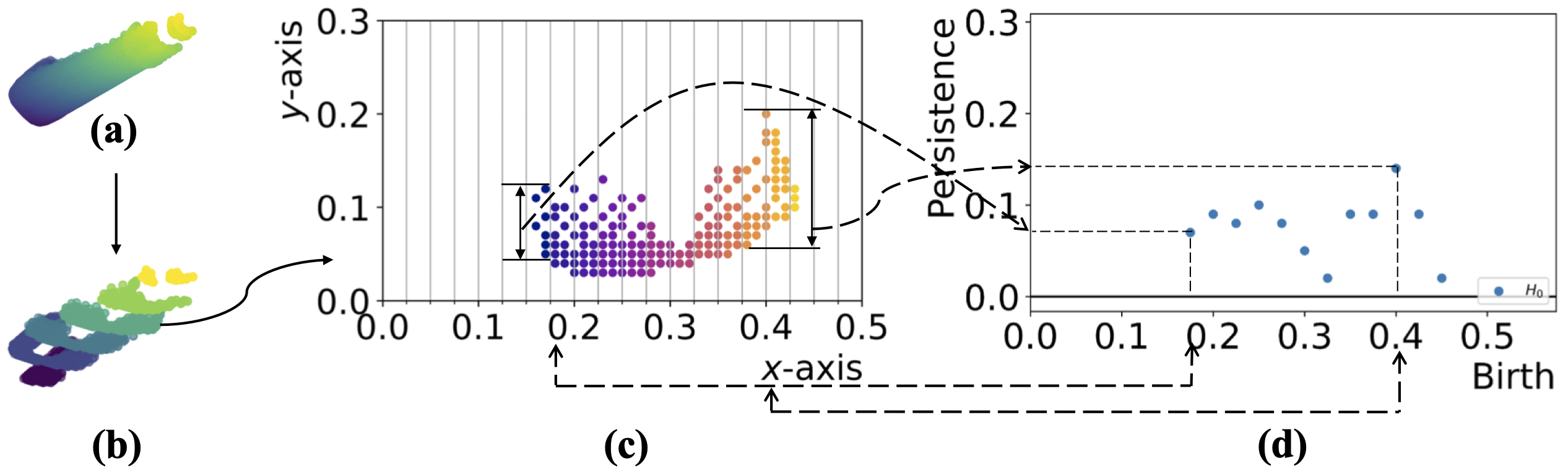}
    \caption{Example of an aligned object point cloud,  $\hat{\mathcal{P}}$ in (a), the slices $\mathcal{S}^0$ to $\mathcal{S}^6$ obtained from it in (b), and a visual mapping between one of its slices $\mathcal{S}^3$ in (c), and the corresponding birth-persistence diagram in (d) showing how the filtration mimics further slicing of $\mathcal{S}^3$ across the $x$-axis.}
    \label{filtration}
\end{figure}

Applying persistent homology to the filtrations gives us a PD for every $\mathcal{S}^i$. We filter the PD such that for each unique value of birth, only the point with the highest persistence is retained. We then compute PIs from the PDs, vectorize and stack them into a single descriptor. Fig. \ref{filtration} illustrates how the descriptor generation works for a sample object.

\begin{figure*}
    \centering
    \includegraphics[width=0.85\textwidth]{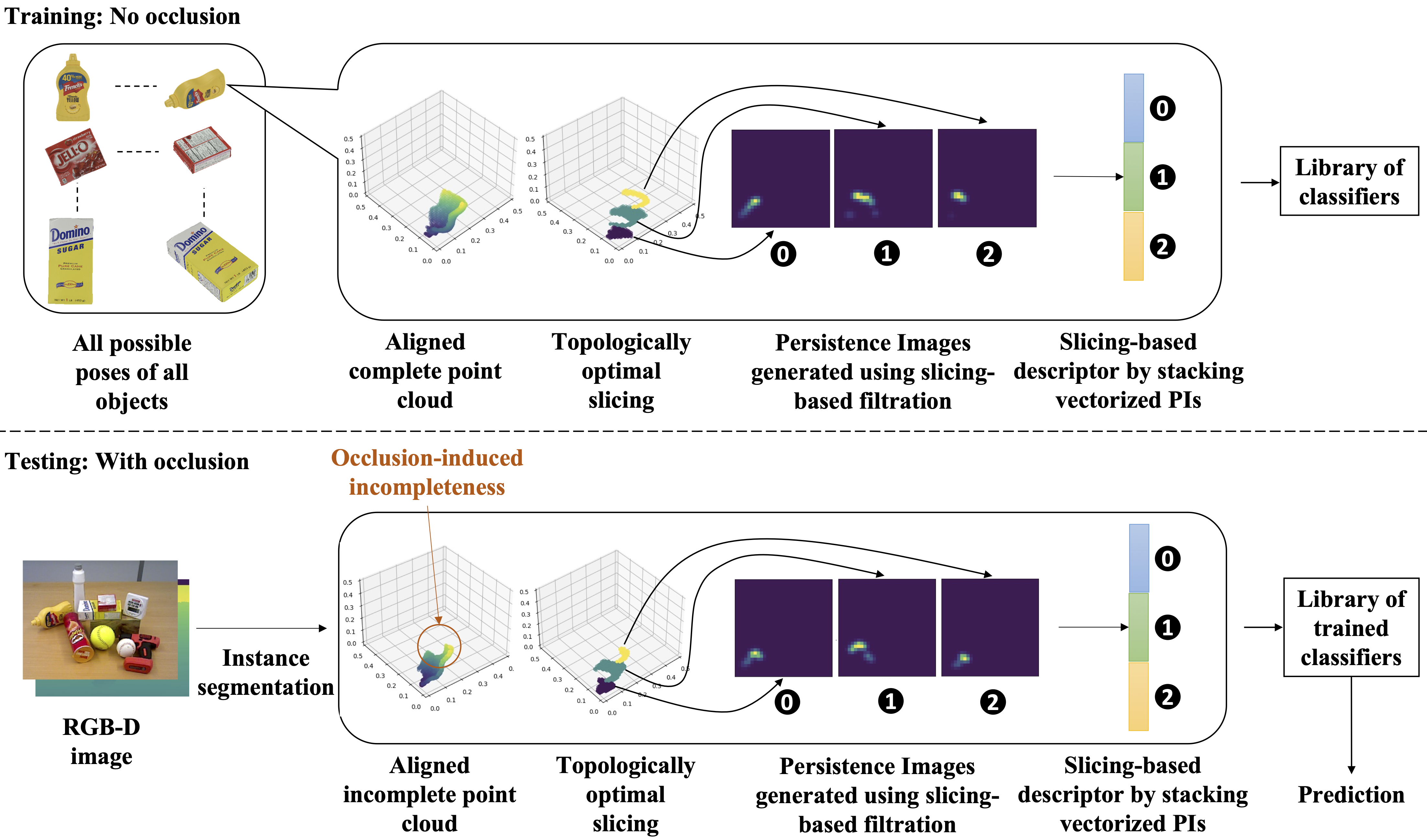}
    \caption{Proposed framework for object recognition using the slicing-based descriptor}
    \label{pipeline}
\end{figure*}

\subsection{Recognition using library of trained models}
\subsubsection{Library generation}

The object point clouds generated from depth images are partial, and the degree of missingness depends on the camera's pose relative to the object. Therefore, we consider all the possible views of all the objects in the training set. Additionally, in cluttered environments, the occlusion of objects causes further incompleteness in the partial point clouds. However, all possible occlusion scenarios cannot be comprehensively captured through data. Therefore, we do not include the point clouds of the occluded objects in the training set. Instead, we perform recognition of partially occluded objects based on the principle of object unity.
From here on, we refer to the partial point clouds of the unoccluded objects as complete object point clouds, and refer to the partial point clouds of occluded objects as incomplete object point clouds. We observe from Fig. \ref{pipeline} that the generated PIs for a mustard bottle in the presence and absence of occlusion have similarities; at test time, the PIs for the slices unaffected by occlusion are similar to the PIs of the corresponding slices of the complete point cloud at train time. We use such similarity to construct a library of trained classifiers that facilitates object unity-based recognition. 

Our training set consists of synthetic point clouds that are all to scale. We divide them into three groups based on the proximity of the viewpoints to the three main orthographic views\footnote{For every object, the orthographic views whose projections have the largest and smallest bounding box areas (among the three views) are termed as the front view and top view, respectively.}. We call these the front, side, and top sets. We then align all the point clouds as described previously. 
Note that we perform data augmentation by mirroring the point clouds across the coordinate axes (in place) before completing the final rotation by $\alpha$. For every set, we begin by computing the slicing-based descriptor of every object point cloud only considering its first slice. We train a classification model (e.g., SVM) using the descriptors. We then consider the first two slices of every object point cloud. We compute the slicing-based descriptor accordingly and train a separate classification model. We continue this procedure until the last slice of the largest object has been considered. As the number of slices differs from object to object, we appropriately pad the descriptor before training any model to ensure that input vectors are all of the same size.


\subsubsection{Using the library at test time}
To recognize $\hat{\mathcal{P}}$, we first obtain the areas of the three orthogonal faces of its minimal-volume bounding box. We assume that the camera pose is known, as is typical for mobile robots. We use it (after necessary transformations) to identify which of the three faces is in direct view. For simplicity, we call this the viewed face. We also use the camera's distance from the center of $\hat{\mathcal{P}}$ to ensure $\hat{\mathcal{P}}$'s scale during descriptor computation is similar to the scale used for training. We then compare the area of the viewed face with the areas of the other two faces and use area-based heuristics to choose the model set(s) from the library for recognition. In some cases, we also use heuristics based on the curvature flow (equivalent to optical flow) of the surfaces corresponding to the faces.
\begin{figure*}[ht!]
    \centering
    \includegraphics[width=0.85\textwidth]{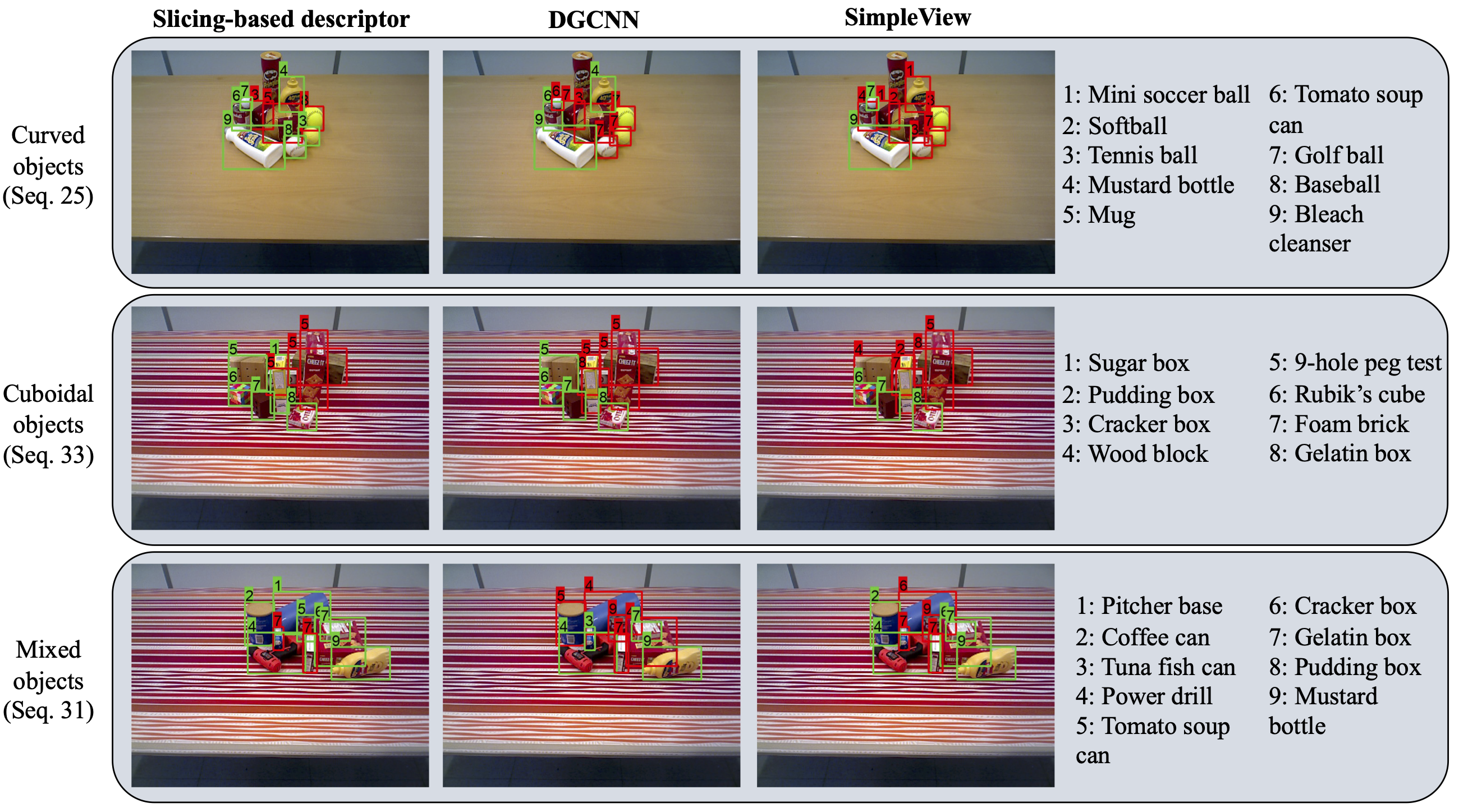}
    \caption{Recognition performance of the proposed method, DGCNN, and SimpleView on sample images from three different types of sequences from the OCID dataset (green boxes indicate correct recognition and red boxes indicate incorrect recognition)}
    \label{resultsfigure}
\end{figure*}

Next, we use the depth value and the segmentation labels of the pixels surrounding the object corresponding to $\hat{\mathcal{P}}$ to identify if the object is occluded. If the object is occluded, $\hat{\mathcal{P}}$ has one or more missing slices. 
In this case, we ensure $\hat{\mathcal{P}}$'s alignment is such that the first slice on the occluded end of the object is not the first slice during descriptor computation. We then identify the number of slices (say $n_s$) in $\hat{\mathcal{P}}$. Then, from the chosen model set(s), we perform recognition using the model(s) that considers only $n_s$ slices. 
If the object is not occluded, we use the model(s) 
that considers the highest number of slices. 
Before using the model(s), $\hat{\mathcal{P}}$'s descriptor is appropriately padded if required. For the case when multiple models are used, the prediction with the highest probability is chosen.

\vspace{-0.5mm}

\section{Experiments and Results}

We use the YCB10 subset of the OCID dataset \cite{suchi2019easylabel} for our experiments. It consists of real-world RGB-D images and instance segmentation maps for sequences of increasingly cluttered scenes with up to ten objects. The sequences are divided into three types - cuboidal (all the objects have sharp edges), curved (all the objects have smooth curved surfaces), and mixed (both cuboidal and curved objects are present). We consider nine different sequences from the subset, three of each type for performance evaluation. We use the Panda3D \cite{panda3d_2018} framework and object meshes from \cite{calli2015benchmarking} to obtain synthetic depth images of the objects for library generation.
We set $\alpha = 45$\degree, $\sigma_1 = 0.1$, and $\sigma_2 = 0.025$ to obtain suitable descriptors. 
Using these descriptors, we generate a separate library of SVMs trained with Platt scaling for every sequence. We perform five-fold cross validation and compare the performance of our method against two state-of-the-art point cloud classification methods, namely, DGCNN\cite{wang2019dynamic} and SimpleView \cite{goyal2021revisiting}, using the implementations provided by the latter. Fig. \ref{resultsfigure} shows the predictions on sample images from three different sequences.

Table \ref{resultstable} shows that our method outperforms DGCNN and SimpleView on all the test sequences. In the case of sequences with curved objects, our method is better at distinguishing between objects with similar geometry, such as the tennis ball, the golf ball, and the baseball (see Seq. 25 in Fig. \ref{resultsfigure}). In the case of sequences with cuboidal objects, the overall performance of all the three methods is better than that for the curved objects. 
We believe that this trend is observed because the objects are more distinguishable (even though they are all of the same cuboidal geometry) due to larger variations in dimensions than the curved objects. Our method outperforms DGCNN and SimpleView in this case too, albeit with a comparatively smaller margin. Our method achieves an even better performance in the case of mixed objects when the object geometries vary considerably. 
The performance of the other methods also improves, but not enough to outperform our method. As shown in Fig. \ref{resultsfigure}, only our method correctly identifies the relatively heavily occluded objects in Seq. 31, i.e., the pitcher base and the tomato soup can. These results indicate that our slicing-based descriptor has more discriminative power than the other learned representations, 
especially in the presence of occlusion. We note that our method does face difficulty in certain occlusion scenarios. For example, all the methods are unable to recognize the objects when their centers are occluded (e.g., wood block in Seq. 33 from Fig. \ref{resultsfigure}).

\begin{table}[]
\centering
\caption{Comparison of mean recognition accuracy (over all the YCB10 object classes in \%) 
on OCID dataset sequences}
\label{resultstable}
\resizebox{\columnwidth}{!}{%
\begin{tabular}{@{}ccccc@{}}
\toprule
Sequence type           & Sequence ID & Slicing-descriptor + SVM & DGCNN & SimpleView \\ \midrule
\multirow{3}{*}{Curved} & Seq. 25      & \textbf{61.41$\pm$0.22}                    & 32.21$\pm$1.48 & 32.29$\pm$3.84      \\ \cmidrule(l){2-5} 
                        & Seq. 26      & \textbf{61.85$\pm$0.74}                    & 42.44$\pm$3.74 & 56.33$\pm$2.54      \\ \cmidrule(l){2-5} 
                        & Seq. 35      & \textbf{55.78$\pm$0.48}                    & 24.81$\pm$2.96 & 19.15$\pm$2.63      \\ \midrule
\multirow{3}{*}{Cuboidal} & Seq. 23      & \textbf{71.68$\pm$0.36}                    & 48.10$\pm$1.96 & 61.08$\pm$0.66      \\ \cmidrule(l){2-5} 
                        & Seq. 24      & \textbf{70.22$\pm$0.26}                    & 42.18$\pm$0.57 & \textbf{68.08$\pm$1.64}      \\ \cmidrule(l){2-5} 
                        & Seq. 33      & \textbf{58.17$\pm$0.25}                    & 43.38$\pm$1.25 & 39.94$\pm$1.08      \\ \midrule
\multirow{3}{*}{Mixed}  & Seq. 22      & \textbf{77.64$\pm$0.00}                    & 57.44$\pm$3.86 & 61.66$\pm$4.12      \\ \cmidrule(l){2-5} 
                        & Seq. 31      & \textbf{67.03$\pm$0.86}                    & 49.30$\pm$3.07 & \textbf{65.59$\pm$3.32}      \\ \cmidrule(l){2-5} 
                        & Seq. 32      & \textbf{70.46$\pm$0.00}                    & 47.13$\pm$1.71 & 60.31$\pm$4.12      \\ \bottomrule
\end{tabular}%
}
\end{table}

\section{Conclusions}

In this work, we present a new slicing-based topological descriptor for recognizing occluded objects in unseen indoor environments. We construct slicing-style filtrations of simplicial complexes from the 
object's point cloud to obtain the descriptor. Our 
approach ensures similarities between the descriptors of the occluded and the corresponding unoccluded objects. We use this similarity to perform recognition based on the principle of object unity using a library of trained models. Comparisons with two state-of-the-art point cloud classification methods, DGCNN and SimpleView, show that our method has the best performance in all the test sequences of a benchmark RGB-D dataset. In the future, we plan to evaluate the performance of our method by replacing SVMs with other classifiers and incorporate object appearance information to further improve recognition performance. 

\addtolength{\textheight}{-10.5 cm}   








\bibliography{references}

\begin{thebibliography}{10}
\providecommand{\url}[1]{#1}
\csname url@rmstyle\endcsname
\providecommand{\newblock}{\relax}
\providecommand{\bibinfo}[2]{#2}
\providecommand\BIBentrySTDinterwordspacing{\spaceskip=0pt\relax}
\providecommand\BIBentryALTinterwordstretchfactor{4}
\providecommand\BIBentryALTinterwordspacing{\spaceskip=\fontdimen2\font plus
\BIBentryALTinterwordstretchfactor\fontdimen3\font minus
  \fontdimen4\font\relax}
\providecommand\BIBforeignlanguage[2]{{%
\expandafter\ifx\csname l@#1\endcsname\relax
\typeout{** WARNING: IEEEtran.bst: No hyphenation pattern has been}%
\typeout{** loaded for the language `#1'. Using the pattern for}%
\typeout{** the default language instead.}%
\else
\language=\csname l@#1\endcsname
\fi
#2}}

\bibitem{chen2018domain}
Y.~Chen, W.~Li, C.~Sakaridis, D.~Dai, and L.~Van~Gool, ``Domain {A}daptive
  {F}aster {R}-{CNN} for object detection in the wild,'' in \emph{IEEE Conf.
  Comput. Vis. Pattern Recognit.}, 2018, pp. 3339--3348.

\bibitem{he2019multi}
Z.~He and L.~Zhang, ``Multi-{A}dversarial {F}aster-{RCNN} for unrestricted
  object detection,'' in \emph{IEEE Int. Conf. Comput. Vis.}, 2019, pp.
  6668--6677.

\bibitem{saito2019strong}
K.~Saito, Y.~Ushiku, T.~Harada, and K.~Saenko, ``Strong-weak distribution
  alignment for adaptive object detection,'' in \emph{IEEE Conf. Comput. Vis.
  Pattern Recognit.}, 2019, pp. 6956--6965.

\bibitem{kim2019diversify}
T.~Kim, M.~Jeong, S.~Kim, S.~Choi, and C.~Kim, ``Diversify and match: {A}
  domain adaptive representation learning paradigm for object detection,'' in
  \emph{IEEE Conf. Comput. Vis. Pattern Recognit.}, 2019, pp. 12\,456--12\,465.

\bibitem{samani2021visual}
E.~U. Samani, X.~Yang, and A.~G. Banerjee, ``Visual object recognition in
  indoor environments using topologically persistent features,'' \emph{IEEE
  Rob. Autom. Lett.}, vol.~6, no.~4, pp. 7509--7516, 2021.

\bibitem{rusu2009detecting}
R.~B. Rusu, A.~Holzbach, M.~Beetz, and G.~Bradski, ``Detecting and segmenting
  objects for mobile manipulation,'' in \emph{IEEE Int. Conf. Comput. Vis.
  Workshops}.\hskip 1em plus 0.5em minus 0.4em\relax IEEE, 2009, pp. 47--54.

\bibitem{marton2010hierarchical}
Z.-C. Marton, D.~Pangercic, R.~B. Rusu, A.~Holzbach, and M.~Beetz,
  ``Hierarchical object geometric categorization and appearance classification
  for mobile manipulation,'' in \emph{IEEE Int. Conf. Humanoid Robot.}\hskip
  1em plus 0.5em minus 0.4em\relax IEEE, 2010, pp. 365--370.

\bibitem{aldoma2012our}
A.~Aldoma, F.~Tombari, R.~B. Rusu, and M.~Vincze, ``{OUR-CVFH}--oriented,
  unique and repeatable clustered viewpoint feature histogram for object
  recognition and 6{DOF} pose estimation,'' in \emph{Joint DAGM (German
  Association for Pattern Recognition) and OAGM Symposium}.\hskip 1em plus
  0.5em minus 0.4em\relax Springer, 2012, pp. 113--122.

\bibitem{wohlkinger2011ensemble}
W.~Wohlkinger and M.~Vincze, ``Ensemble of shape functions for 3d object
  classification,'' in \emph{IEEE Int. Conf. Rob. Biomimetics}.\hskip 1em plus
  0.5em minus 0.4em\relax IEEE, 2011, pp. 2987--2992.

\bibitem{aldoma2011cad}
A.~Aldoma, M.~Vincze, N.~Blodow, D.~Gossow, S.~Gedikli, R.~B. Rusu, and
  G.~Bradski, ``{CAD}-model recognition and 6{DOF} pose estimation using 3d
  cues,'' in \emph{IEEE Int. Conf. Comput. Vis. Workshops}.\hskip 1em plus
  0.5em minus 0.4em\relax IEEE, 2011, pp. 585--592.

\bibitem{guo20143d}
Y.~Guo, M.~Bennamoun, F.~Sohel, M.~Lu, and J.~Wan, ``3{D} object recognition in
  cluttered scenes with local surface features: A survey,'' \emph{IEEE Trans.
  Pattern Anal. Mach. Intell.}, vol.~36, no.~11, pp. 2270--2287, 2014.

\bibitem{biasotti2016recent}
S.~Biasotti, A.~Cerri, A.~Bronstein, and M.~Bronstein, ``Recent trends,
  applications, and perspectives in 3{D} shape similarity assessment,'' in
  \emph{Computer graphics forum}, vol.~35, no.~6.\hskip 1em plus 0.5em minus
  0.4em\relax Wiley Online Library, 2016, pp. 87--119.

\bibitem{beksi2018signature}
W.~J. Beksi and N.~Papanikolopoulos, ``Signature of topologically persistent
  points for 3{D} point cloud description,'' in \emph{IEEE Int. Conf. Rob.
  Autom.}, 2018, pp. 3229--3234.

\bibitem{qi2017pointnet}
C.~R. Qi, H.~Su, K.~Mo, and L.~J. Guibas, ``Point{N}et: {D}eep learning on
  point sets for 3{D} classification and segmentation,'' in \emph{IEEE Conf.
  Comput. Vis. Pattern Recognit.}, 2017, pp. 652--660.

\bibitem{qi2017pointnet++}
C.~R. Qi, L.~Yi, H.~Su, and L.~J. Guibas, ``Point{N}et++: {D}eep hierarchical
  feature learning on point sets in a metric space,'' \emph{Adv. Neural Inform.
  Process. Syst.}, vol.~30, 2017.

\bibitem{wang2019dynamic}
Y.~Wang, Y.~Sun, Z.~Liu, S.~E. Sarma, M.~M. Bronstein, and J.~M. Solomon,
  ``Dynamic {G}raph {CNN} for learning on point clouds,'' \emph{ACM Trans.
  Graphics}, vol.~38, no.~5, pp. 1--12, 2019.

\bibitem{goyal2021revisiting}
A.~Goyal, H.~Law, B.~Liu, A.~Newell, and J.~Deng, ``Revisiting point cloud
  shape classification with a simple and effective baseline,'' in \emph{Int.
  Conf. Mach. Learn.}, 2021, pp. 3809--3820.

\bibitem{ren2022benchmarking}
J.~Ren, L.~Pan, and Z.~Liu, ``Benchmarking and analyzing point cloud
  classification under corruptions,'' \emph{arXiv preprint arXiv:2202.03377},
  2022.

\bibitem{rapp2015handbook}
B.~Rapp, \emph{Handbook of cognitive neuropsychology: {W}hat deficits reveal
  about the human mind}.\hskip 1em plus 0.5em minus 0.4em\relax Psychology
  Press, 2015.

\bibitem{goldstein2016sensation}
E.~B. Goldstein and J.~Brockmole, \emph{Sensation and perception}.\hskip 1em
  plus 0.5em minus 0.4em\relax Cengage Learning, 2016.

\bibitem{adams2017persistence}
H.~Adams, T.~Emerson, M.~Kirby, R.~Neville, C.~Peterson, P.~Shipman,
  S.~Chepushtanova, E.~Hanson, F.~Motta, and L.~Ziegelmeier, ``Persistence
  images: {A} stable vector representation of persistent homology,'' \emph{J.
  Mach. Learn. Res.}, vol.~18, no.~1, pp. 218--252, 2017.

\bibitem{o1985finding}
J.~O'Rourke, ``Finding minimal enclosing boxes,'' \emph{Int. J. Comput. \&
  Inform. Sci.}, vol.~14, no.~3, pp. 183--199, 1985.

\bibitem{suchi2019easylabel}
M.~Suchi, T.~Patten, D.~Fischinger, and M.~Vincze, ``Easy{L}abel: {A}
  semi-automatic pixel-wise object annotation tool for creating robotic {RGB-D}
  datasets,'' in \emph{IEEE Int. Conf. Rob. Autom.}, 2019, pp. 6678--6684.

\bibitem{panda3d_2018}
\BIBentryALTinterwordspacing
``Panda3{D}: {O}pen source framework for 3{D} rendering \& {G}ames,'' May 2018.
  [Online]. Available: \url{https://www.panda3d.org/}
\BIBentrySTDinterwordspacing

\bibitem{calli2015benchmarking}
B.~Calli, A.~Walsman, A.~Singh, S.~Srinivasa, P.~Abbeel, and A.~M. Dollar,
  ``Benchmarking in manipulation research: {U}sing the {Y}ale-{CMU}-{B}erkeley
  object and model set,'' \emph{IEEE Rob. Autom. Mag.}, vol.~22, no.~3, pp.
  36--52, 2015.

\end{thebibliography}

\end{document}